\documentclass{article}

\usepackage[final]{neurips_2025}

\usepackage{subcaption}
\usepackage{amsmath}
\usepackage[utf8]{inputenc} 
\usepackage[T1]{fontenc}    
\usepackage{hyperref}       
\usepackage{url}            
\usepackage{booktabs}       
\usepackage{amsfonts}       
\usepackage{nicefrac}       
\usepackage{microtype}      
\usepackage{xcolor}         
\usepackage{graphicx}
\usepackage{comment}
\title{Scalable Bayesian Network Structure Learning Using Tsetlin Machine to Constrain the Search Space}

\author{%
  Kunal Ganesh Dumbre \\
  Department of ICT\\
  University of Agder\\
  Grimstad, Norway \\
  \texttt{kunalgd@uia.no}\\
  \And
  Lei Jiao\\
  Department of ICT\\
  University of Agder\\
  Grimstad, Norway \\
  \texttt{lei.jiao@uia.no}\\
  \And
  Ole-Christoffer Granmo\\
  Department of ICT\\
  University of Agder\\
  Grimstad, Norway \\
  \texttt{ole.granmo@uia.no}\\
}

\begin{document}

\maketitle

\begin{abstract}
  The PC algorithm is a widely used method in causal inference for learning the structure of Bayesian networks. Despite its popularity, the PC algorithm suffers from significant time complexity, particularly as the size of the dataset increases, which limits its applicability in large-scale real-world problems. In this study, we propose a novel approach that utilises the Tsetlin Machine (TM) to construct Bayesian structures more efficiently. Our method leverages the most significant literals extracted from the TM and performs conditional independence (CI) tests on these selected literals instead of the full set of variables, resulting in a considerable reduction in computational time. We implemented our approach and compared it with various state-of-the-art methods. Our evaluation includes categorical datasets from the bnlearn repository, such as \textit{Munin1}, \textit{Hepar2}. The findings indicate that the proposed TM-based method not only reduces computational complexity but also maintains competitive accuracy in causal discovery, making it a viable alternative to traditional PC algorithm implementations by offering improved efficiency without compromising performance.

\end{abstract}

\section{Introduction}\label{Introduction}
Bayesian Networks (BNs) are a type of probabilistic graphical model that use directed acyclic graphs~(DAGs) to efficiently represent random variables along with their conditional dependencies. The graph-based structure of BNs makes them particularly effective for modeling uncertain knowledge and performing inference \citep{LABORDA2024111840}. BNs have proven useful across numerous real-world domains and applications \citep{KYRIMI2021102108}\citep{6722f42e4d86470387f1402dcf5c541f}

The process of constructing a BN from data typically involves two primary approaches: constraint-based methods and score-based methods. In this work, we focus on constraint-based methods, which rely on Conditional Independence (CI) tests to identify the structure of the network. The PC algorithm is a well-known example of this approach and can be divided into two main variants: Original PC (PCO) \citep{10.7551/mitpress/1754.001.0001} and PC-Stable (PCS) \citep{10.5555/2627435.2750365}. The PCO rely on CI tests to identify the structure of the network. The PCO first constructs a fully connected graph and then systematically removes edges based on CI tests, thereby uncovering the underlying structure of the data. However, the PCO is known to be sensitive to the order in which variables are processed. To mitigate this, the PCS was introduced as a modified constraint-based approach that eliminates order dependence by enforcing a consistent edge-removal strategy across all CI tests. Both of these algorithms, however, face exponential time complexity as the size of the dataset increases. The time required to compute the CI tests grows exponentially with the number of variables, making these methods computationally expensive for large datasets \citep{7513439}. As such, scalability remains a significant challenge for constructing BNs from large, high-dimensional datasets.

To reduce the computational burden of CI testing, several techniques have been proposed that make use of the Markov Blanket (MB) concept to minimise the number of tests required. The MB of a variable (i.e., a feature) in a BN consists of the variable’s parents, children, and spouses (other parents of the variable’s children). MB discovery plays an essential role in scalable BN structure learning and optimal feature selection. For example, by finding the MB of each variable in a dataset, we can use the discovered MBs as constraints to develop efficient and effective local-to-global BN structure learning algorithms.
Margaritis and Thrun \citep{NIPS1999_5d79099f} introduced the first MB discovery method, the Grow-Shrink (GS) algorithm, which operates in two stages: a growing phase that identifies potential MB candidates and a shrinking phase that removes false positives. The Incremental Association Markov Blanket (IAMB) algorithm later improved GS by selecting variables with the highest conditional association rather than random selection. Several variants, including inter-IAMB, IAMBnPC, and Fast-IAMB \citep{1565788}, enhanced efficiency but required exponentially more samples with larger MBs and struggled to distinguish between parents-children (PC) and spouse (SP) nodes. To overcome these issues, algorithm HITON-MB \citep{article-hitonmb} adopted divide-and-conquer strategies to separately identify PC and SP, although they were later found incorrect under the faithfulness assumption \citep{PENA2007211}.

The Parents-and-Children-based Markov Blanket algorithm (PCMB) algorithm \citep{PENA2007211} was the first to ensure correctness under faithfulness by applying a symmetry constraint, though with high computational cost. The Simultaneous Markov Blanket (STMB) \citep{7440796} algorithm addressed this by first finding the PC set and then identifying SPs, improving speed but slightly reducing accuracy due to conditioning on full sets. The most recent Fast Shrinking parents-children learning for Markov blanket-based feature selection (FSMB) \citep{article-fastsmb} algorithm optimizes this process by detecting PC first, discovering and validating SPs concurrently, and immediately removing false positives through a secondary screening step. By integrating SP validation with PC rechecking and employing the AdjV algorithm for PC identification, FSMB achieves faster and more accurate MB discovery.

The aim of this paper is to develop a new Parent-childrens algorithm that improves efficiency while maintaining accuracy. The main contributions are summarised below:
 (1) We present a new method that employs the TM for constructing causal structures more efficiently than traditional methods.   
    (2) By performing CI tests on important literals, our approach significantly reduces computational time compared to the other algorithms without sacrificing performance, making it suitable for high-dimensional datasets. We implemented our method and compared its performance with the PCS, PCO, HITON-PC, FSMB-PC.
    
    

The rest of the paper is organized as follows. Section ~\ref{Background} presents the conceptual background and key definitions related to Bayesian Networks, Conditional Independence, and Markov Blankets. Section~\ref{TM_overview} provides an overview of the TM, a rule-based algorithm used in our approach, explaining its principles. Section~\ref{Proposed_method} introduces our technique designed to reduce the time complexity. We provide a detailed explanation of our method. Section ~\ref{assumptions} discusses the underlying assumptions, while Section ~\ref{complexity} analyzes the computational complexity of the method. Section~\ref{Experiments} presents the experimental evaluation of our technique, including the setup, datasets, and performance metrics used to assess its effectiveness. We compare our results with those of the state-of-the-art approaches to demonstrate the improvements achieved before we conclude in Section~\ref{Conclusion}. 

\section{Conceptual background} \label{Background}
In this section, we introduce the theoretical concepts and notations that form the foundation of our study. We first formalize key definitions related to BNs, CI, and MBs. These concepts are central to understanding both traditional constraint-based approaches and our proposed Tsetlin Machine--based framework.

\subsection{Notations, definitions, and formalization}

Let $U$ denote a finite set of random variables. A \textit{Bayesian network (BN)} is defined as a triplet $\langle U, G, P \rangle$, where $G$ is a DAG over $U$, and $P$ is the joint probability distribution that factorizes according to $G$. Specifically, each variable $X \in U$ is associated with a conditional probability distribution given its parents in $G$.

\textbf{Definition 1 (Conditional independence).}  
Two variables $X, Y \in U$ are conditionally independent given a set $Z \subseteq U \setminus \{X,Y\}$, denoted as $X \perp\!\!\!\perp Y \mid Z$, if and only if
\begin{equation}
P(X,Y \mid Z) = P(X \mid Z) \, P(Y \mid Z).
\end{equation}

\textbf{Definition 2 (Faithfulness).}  
A BN $\langle U, G, P \rangle$ is \textit{faithful} if all and only the conditional independencies in $P$ correspond to d-separations in $G$.

\textbf{Definition 3 (V-structure).}  
A triplet of variables $(X, Z, Y)$ forms a v-structure if $X \to Z \leftarrow Y$ and $X$ and $Y$ are non-adjacent.

\textbf{Definition 4 (Markov blanket).}  
For a target variable $X_T \in U$, the Markov blanket $MB_T$ consists of its parents, children, and spouses. It satisfies the property that $X_T$ is conditionally independent of all other variables in $U \setminus (X_T \cup MB_T)$ given $MB_T$.

\textbf{Definition 5 (D-separation).}  
A path between $X$ and $Y$ is \emph{blocked} by $S \subseteq U \setminus \{X, Y\}$ if it contains a non-collider in $S$ or a collider such that neither it nor its descendants are in $S$; $X$ and $Y$ are \emph{d-separated} by $S$ (denoted $X \perp\!\!\perp_G Y \mid S$) when all paths between them are blocked.

Table~\ref{tab:notation} summarizes the notation adopted throughout this paper.

\begin{table}[h]
\centering
\caption{Summary of notations.}
\label{tab:notation}
\begin{tabular}{ll}
\toprule
\textbf{Notation} & \textbf{Meaning} \\
\midrule
$U$ & Set of variables (features) \\
$G$ & Directed acyclic graph (DAG) over $U$ \\
$P$ & Joint probability distribution over $U$ \\
$X, Y$ & Random variables in $U$ \\
$Z, S$ & Conditioning sets within $U$ \\
$X \perp\!\!\!\perp Y \mid Z$ & $X$ is conditionally independent of $Y$ given $Z$ \\
$MB_T$ & Markov blanket of target variable $T$ \\
$PC_T$ & Parents and children of $T$ \\
$SP_T$ & Spouses of $T$ \\
$|U|$ & Total number of variables \\
\bottomrule
\end{tabular}
\end{table}

\section{Tsetlin machine overview} \label{TM_overview}
The TM uses teams of simple Tsetlin Automata (TA) to construct propositional logic-based pattern recognition systems. Unlike traditional models, it relies on efficient bitwise operations and a unique game-theoretic approach to guide learning, avoiding float number calculation. 
This approach enables competitive accuracy and, in some cases, outperforms recent attention-based neural networks \citep{Sharma_Yadav_Granmo_Jiao_2023}. They have been successfully applied in various areas, such as sentiment analysis \citep{Yadav_Jiao_Granmo_Goodwin_2021}, robust interpretation \citep{ijcai2022p616}, image classification \citep{Jeeru_article}, contextual bandit problems \citep{NEURIPS2022_c2d550cf}, and federated learning \citep{qi2025fedtmos}. TMs are hardware-friendly and energy-efficient on dedicated hardware \citep{tunheim2025alldigital86njframe65nmtsetlin}, making them ideal for IoT devices and edge computing, where efficiency and low power are crucial. In terms of performance, convergence studies show that TMs reliably reach accurate boolean operations after training \citep{9881240}\citep{9445039}. The details of TM can be found in \citep{granmo2021tsetlinmachinegame} and we summarize briefly the operational concept below.

\subsection{Tsetlin machine concept}

Consider a Boolean input feature vector \( \mathbf{x} = [x_1, x_2, \ldots, x_o] \in \{0,1\}^o \), and the TM constructs a set of literals \( L = \{x_1, \ldots, x_o, \neg x_1, \ldots, \neg x_o\} \), which combines both the original features and their negations of the input. These literals serve as the basic building blocks for constructing \textit{conjunctive clauses}, using logical AND operations over selected literals, that capture sub-patterns in the data.

To decide if a literal is included as a part of a clause or not, each literal corresponds to a TA, which is a two-action learning automaton with $2N$ states. States $1$ to $N$ correspond to the \textit{Exclude} action, while states $N + 1$ to $2N$ correspond to \textit{Include}. TAs receive feedback—Reward, Penalty, or Inaction—based on the correctness of classifications during training. Rewards strengthen the current action, while penalties move the TA toward the center, encouraging a switch. This mechanism allows the TM to adaptively learn meaningful clauses during training.

The TM uses a total of \( n \) clauses per class, where \( n \) is a user-defined hyperparameter. Half of the clauses are assigned positive polarity ($+$), while the remaining half are assigned negative polarity ($-$).  Each clause $C_j$ in the TM selects a subset of these literals, denoted as $L_j \subseteq L$. Here, $j$ indexes the clauses, and $L_j$ refers to the set of literals chosen by clause $C_j$ to form its conjunctive expression.

For a binary classification task, each clause outputs 1 if the conjunction of its selected literals evaluates to true, and 0 otherwise. The TM aggregates clause outputs into a signed sum :
\begin{equation}
s(\mathbf{x}) = \sum_{j=1}^{n/2} C_j^+(\mathbf{x}) - \sum_{j=1}^{n/2} C_j^-(\mathbf{x}).
\label{eq:vote_sum}
\end{equation}
In Eq.~(\ref{eq:vote_sum}),  \( C_j^+ \) and \( C_j^- \) denote the outputs of positive and negative clauses, respectively. Clearly,  \( C_j^+ \) vote for the class while  \( C_j^- \) vote against the class. The final prediction is determined by applying a unit step function as shown in Eq.~(\ref{eq:unit_step}) :
\begin{equation}
\hat{y} = u(s(\mathbf{x})).
\label{eq:unit_step}
\end{equation}

In the multi-class classification setting, the TM maintains a distinct set of \( n \) clauses for each class \( i \in \{1, \ldots, m\} \), again split evenly between positive and negative polarity. Each class independently computes a signed sum of its clause outputs. The final class prediction is the one with the highest total as shown in Eq.~(\ref{eq:multi_class}):
\begin{equation}
\hat{y} = \operatorname*{arg}\operatorname*{max}_{i} \left( s_i(\mathbf{x}) \right).
\label{eq:multi_class}
\end{equation}

\subsection{Tsetlin machine learning}
The TM uses reinforcement learning, where each clause receives feedback and passes it to its TA. There are two types of feedback: Type I and Type II. Type I  encourages a clause to evaluate to 1 for the current input. If the clause already evaluates to 1, it is refined by adding more literals. Type II drives the clause to evaluate to 0. These feedback mechanisms adjust the states of the TA, modifying the literals in the clauses in a way that over time improves the classification accuracy.

Training begins by randomly initializing the states of the individual TA, resulting in clauses containing randomly selected literals. The TM is fed with training data, one example $(\mathbf{x}, y)$ at a time. For an input $\mathbf{x}$ of class $y = 0$, the objective is to ensure that the signed sum of the clauses, denoted as $s(\mathbf{x})$, becomes negative (see Eq.~(\ref{eq:vote_sum})). If this doesn't happen, Type I feedback is provided to some negative clauses, and Type II feedback is given to some positive clauses. Conversely, for an input $\mathbf{x}$ of class $y = 1$, the goal is to make $s(\mathbf{x})$ non-negative. If the signed sum is negative, Type I feedback is given to some positive clauses, and Type II feedback to some negative clauses.

To introduce an ensemble effect, for each training example $(\mathbf{x}, y)$, feedback is given to a random selection of clauses based on a hyper-parameter $T$ (a target value). The TM aims to adjust the signed sum $s(\mathbf{x})$ such that $s(\mathbf{x})$ reaches $-T$ for class $y = 0$, and $s(\mathbf{x})$ reaches $T$ for class $y = 1$. The ensemble effect is achieved by giving Type I and Type II feedback to clauses randomly.

First, clamp $s(\mathbf{x})$ between $-T$ and $T$:
\begin{equation}
c(\mathbf{x}) = \text{clamp}(s(\mathbf{x}), -T, T).
\label{eq:clamp_eq}
\end{equation}

Next, each clause receives feedback with probability $p_c(\mathbf{x})$, proportional to the difference between the clamped sum and the target $T$:
\begin{equation}
p_c(\mathbf{x}) = 
\begin{cases} 
\frac{T + c(\mathbf{x})}{2T}, & \text{if } y = 0, \\
\frac{T - c(\mathbf{x})}{2T}, & \text{if } y = 1.
\end{cases}
\label{eq:feedback_probability}
\end{equation}

This random clause selection encourages the clauses to spread across diverse sub-patterns, rather than clustering on a few. This occurs because the feedback frequency is proportional to the difference between the clamped sum $c(\mathbf{x})$ and the target $\pm T$. As $s(\mathbf{x})$ approaches the target, feedback decreases and stops once the target is reached. This ensures that only a subset of clauses is activated to recognize each frequent sub-pattern, maintaining diversity.

\subsection{Weighted Tsetlin machine}
Weighted Tsetlin machine (WTM) was introduced by \citep{phoulady2020weightedtsetlinmachinecompressed} which extends the standard TM by assigning a weight to each clause instead of repeating similar clauses to increase their impact. This allows for more compact models and finer control over clause influence. Weights are initially set to 1 and updated through modified feedback: Type I feedback increases weights for correct predictions, while Type II feedback decreases weights to reduce false positives. Classification is then performed using a weighted majority vote:
\begin{equation}
s(\mathbf{x}) = \sum_{j=1}^{n/2} w_j^+ C_j^+(\mathbf{x}) - \sum_{j=1}^{n/2} w_j^- C_j^-(\mathbf{x}).
\label{eq:weight_tm}
\end{equation}

\section{Proposed method} \label{Proposed_method}

This section describes our approach for constructing a BN by identifying the most influential variables for each target variable $X_T \in U$ using a WTM, followed by CI testing to analyze their relationships.

\subsection{Identifying important variables using WTM}

\begin{figure}
  \centering
  \includegraphics[width=1\linewidth]{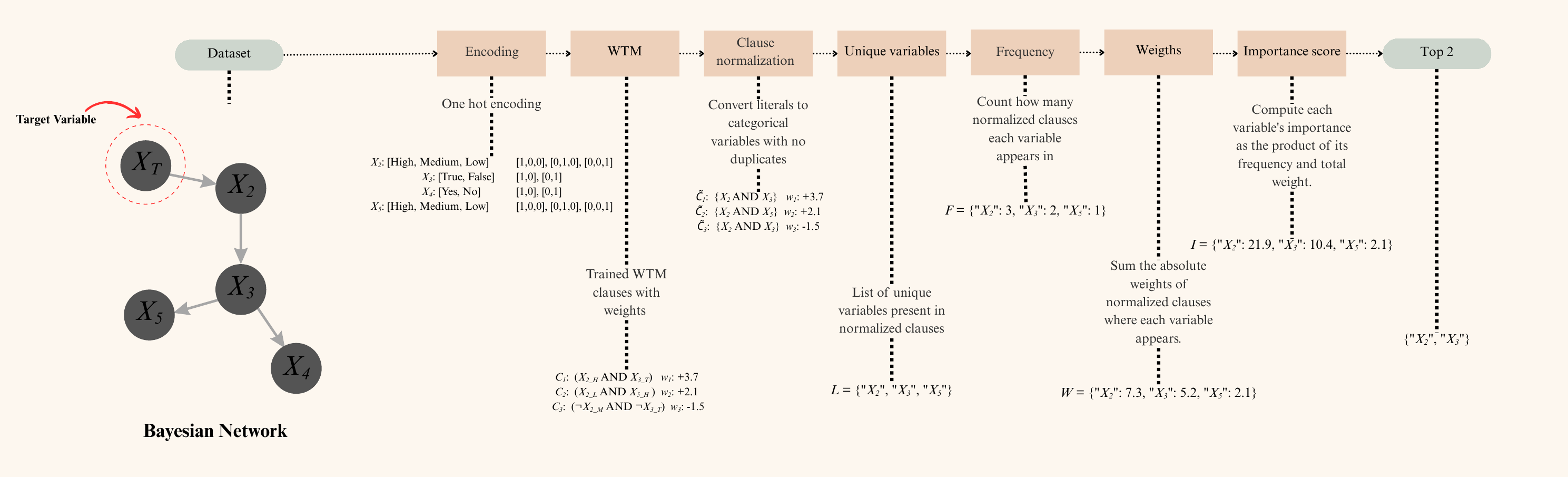} 
  \caption{Workflow for identifying important variables using WTM.}
  \label{fig:top_2}
\end{figure}

We extract important variables using the following 7 steps, highlighted in Fig.~\ref{fig:top_2}. 

\subsubsection{Step 1: Encoding}
The input dataset is fully binarized using one-hot encoding. That is, a categorical variable $X_k$ with $m$ possible values is transformed into $m$ binary variables: $X_{k,v_1}, X_{k,v_2}, \ldots, X_{k,v_m}$. Each binary variable represents one category of $X_k$.

\subsubsection{Step 2: Weigthed Tsetlin machine}
We train a WTM on the binarized data to predict the target variable $X_T$. Each clause $C_j$ in the trained model is a combination of binary literals.

\subsubsection{Step 3: Clause normalization}

Each WTM clause consists of \textit{literals} (i.e., binary variables or their negations). Since these literals originate from one-hot encoded variables, we group them back into their original categorical variables.

Given a learned clause \( C_j \), we define its \textit{normalized clause} \( \tilde{C}_j \) as:
\begin{equation}
  \tilde{C}_j = \{ X_k \mid \exists \, X_{k,v} \text{ or } \neg X_{k,v} \in C_j \}.
\label{eq:nor_clause}
\end{equation}
This transformation ensures that each original variable \( X_k \) appears at most once in \( \tilde{C}_j \), regardless of how many of its binary literals are included.

\subsubsection{Step 4: Unique variables}

Let \( L \) be the set of unique original variables that appear in any normalized clause associated with the target variable \( X_T \).

\subsubsection{Step 5: Frequency}

For each variable \( X_k \in L \), we count how many normalized clauses it appears in:
\begin{equation}
  F(X_k) = \sum_j A(X_k \in \tilde{C}_j),
  \label{eq:nor_frequency}
    \end{equation}
  where \( A(\cdot) \) is the indicator function.

\subsubsection{Step 6: Weights}

Each normalized clause \( \tilde{C}_j \) has an associated weight \( w_j \). For each  \( X_k \), we compute the total absolute weight:
\begin{equation}
  W(X_k) = \sum_{j : X_k \in \tilde{C}_j} |w_j|.
\label{eq:sum_weights}
    \end{equation}

\subsubsection{Step 7: Importance score}

The \textit{importance score} for each variable is defined as:
    \begin{equation}
  I(X_k) = F(X_k) W(X_k).
\label{eq:important_score}
    \end{equation}

We denote the importance score by \(I(X_k)\). Variables are ranked by their importance scores. For each target variable, we select the \textit{top 2 variables} as the most important.

\subsection{Conditional independence testing}

Once the top-2 important variables for a target variable \( X_T \) are identified (e.g., \( X_2 \) and \( X_3 \)), we test whether their influence is direct or mediated by other variables. Let
\[
\mathcal{N}(X_T) = \{X_2, X_3\}, \quad
\mathcal{N}(X_2) = \{X_3, X_5\}, \quad
\mathcal{N}(X_3) = \{X_2, X_4\}.
\]
Here, \(\mathcal{N}(X_i)\) denotes the set of the top-2 important variables (neighbors) for variable \(X_i\).

The CI tests are conducted between the target variable and its important variables. For the target variable \( X_T \), the tests are performed as follows:

\begin{itemize}
    \item For $X_T$ and $X_2$: $X_T \perp\!\!\!\perp X_2 \,\big|\, (\mathcal{N}(X_T) \setminus \{X_2\}) \cup \left( \mathcal{N}(X_2) \setminus \{X_T\} \right)$.

    \item For $X_T$ and $X_3$: $X_T \perp\!\!\!\perp X_3 \,\big|\, (\mathcal{N}(X_T) \setminus \{X_3\}) \cup \left( \mathcal{N}(X_3) \setminus \{X_T\} \right)
    $.
\end{itemize}
The same is performed for other target variables.

\subsection{Constructing the causal graph}
Once the independent and CI tests have been performed, the results are used to construct a BN.

\begin{itemize}
    \item If two variables $X_i, X_j$ are independent ($X_i \perp\!\!\perp X_j$), no direct edge is added between them in $G$.
    \item If $X_i$ and $X_j$ are dependent and remain dependent given any conditioning set $S \subseteq U \setminus \{X_i, X_j\}$, a direct edge is added between them.
    \item If $X_i$ and $X_j$ are dependent, but become conditionally independent given a conditioning set $S$ ($X_i \perp\!\!\perp X_j \mid S$), this suggests that the causal influence is mediated through $S$, no direct edge is added between them
in $G$.
\end{itemize}

Thus, using the results of the CI tests, we can iteratively add edges between variables, resulting in the construction of the BN.

\section{Assumptions} \label{assumptions}

The proposed algorithm is developed based on the following key assumptions:

\begin{enumerate}
    \item \textbf{Causal sufficiency} \\
    It is assumed that all common causes of the observed variables have been measured and included in a dataset
    
    \item \textbf{Sparsity of influences} \\
    Each variable is assumed to be directly influenced by only a few other variables. This sparsity assumption is reflected in the design of the algorithm, where only the top two most important variables are selected for each target. It is assumed that the dominant causal effects are limited in number and that weaker or higher-order dependencies can be ignored without much loss of information.
    
    \item \textbf{Faithfulness} \\
    It is assumed that the observed conditional independencies among variables truly represent the causal independencies in the underlying system. This ensures that the results of the conditional independence tests correctly correspond to actual causal relationships rather than random statistical associations.
    
    \item \textbf{Meaningfulness of WTM weights} \\
    The method assumes that the frequency and weight of literals in WTM clauses provide a reasonable indication of a variable’s influence. Variables appearing more often or with larger absolute weights are assumed to have stronger effects.
    
    \item \textbf{Sufficient sample size} \\
    The dataset is assumed to have enough samples to learn reliable WTM weights and to perform stable conditional independence tests. Insufficient data may lead to unreliable patterns.
\end{enumerate}

\section{Complexity} \label{complexity}

In this section, we analyze the computational complexity of the proposed method for identifying the most influential variables for a single target variable $X_T$. As in previous works, the complexity is expressed in terms of the number of CI tests, since they represent the primary computational cost in most MB discovery algorithms.

For $X_T$, a WTM is trained using all other variables. With $C$ clauses and $E$ epochs over $|U|$ variables, the training complexity is 
\begin{equation}
O(E \, C \, |U|).
\end{equation}
After identifying the top-2 important variables $\mathcal{N}(X_T) = \{X_a, X_b\}$, separate WTM models are trained for each to obtain $\mathcal{N}(X_a)$ and $\mathcal{N}(X_b)$, keeping the total cost at $O(E \, C \, |U|)$ (constants omitted).

CI tests are performed between $X_T$ and each top variable. For $(X_T, X_a)$, the conditioning set
\begin{equation}
S_{Ta} = (\mathcal{N}(X_T) \setminus \{X_a\}) \cup (\mathcal{N}(X_a) \setminus \{X_T\})
\end{equation}
has at most 3 variables, resulting in $2^3 = 8$ tests. Similarly, $(X_T, X_b)$ requires 8 tests, totaling 16 CI tests per target.

Overall, the complexity per target is
\begin{equation}
O(E \, C \, |U|) + O(16) = O(E \, C \, |U|),
\end{equation}
dominated by WTM training and linear in the number of features. The method thus requires a fixed small number of CI tests and linear-time WTM training, reducing computational cost compared to existing MB algorithms.

\section{Experiments}  \label{Experiments}

\subsection{Experimental setup}\label{Experiments_setup}
The proposed TM-based technique was implemented in Python and benchmarked against PCS and PCO from the pgmpy library. We also compared our method's CI test performance with other approaches, including HITON-PC, and FSMB-PC.  Experiments were conducted on a MacBook Pro M3 (2.6 GHz, 16 GB RAM). Hyperparameters by network size are listed in Table~\ref{Hyper-parameters-table}. We used datasets from six benchmark Bayesian networks (Table~\ref{Bayesian-network}), each with 5,000 complete samples across diverse domains. For training the TM, we employed a balanced 80/20 train-test split. However, for performing conditional independence testing, we used the unbalanced full dataset. Evaluation metrics included precision, recall, f1-score, and CI tests. Each algorithm was run ten times, with averaged results. Node differences, Missing edges, Wrong edges, Structure Humming Distance (SHD) were rounded up for comparison and presented in apendix.

\begin{table}[htbp]
  \caption{Hyper-parameters.}
  \label{Hyper-parameters-table}
  \centering
  \begin{tabular}{lcccc}
    \toprule
    \textbf{Network size} & \textbf{Threshold ($T$)} &
    \textbf{Clause ($C$)} &
    \textbf{Specificity ($s$)} & \textbf{Epochs ($E$)} \\
    \midrule
    (Nodes <30)      & 50 & 50 & 5  & 2   \\
    (Nodes 30-50)   & 50 & 50 &5  & 5   \\
    (Nodes >50)     & 50 & 50 &5  & 10  \\
    \bottomrule
  \end{tabular}
\end{table}

\begin{table}[htbp]
  \caption{Bayesian networks used in the experiments.}
  \label{Bayesian-network}
  \centering
  \begin{tabular}{lrrrrrrrr}
    \toprule
    &\scriptsize \textbf{Insurance} & \scriptsize \textbf{Water} & \scriptsize \textbf{Alarm} & \scriptsize \textbf{Barley} & \scriptsize \textbf{Hailfinder} & \scriptsize \textbf{Hepar2} & \scriptsize \textbf{Munin1} \\
    \midrule
    Nodes & 27 & 32 & 37 & 48 & 56 & 70 & 186 \\
    Edges & 52 & 66 & 46 & 84 & 66 & 123 & 273 \\
    Parameters & 1008 & 10083 & 509 & 114005 & 2656 & 1453 & 15622 \\
    \bottomrule
  \end{tabular}
\end{table}

\subsection{Results}\label{Result}

\begin{table}
  \centering
  \caption{Number of CI tests performed by algorithms on different datasets (mean ± standard deviation)}
  \label{tab:ci-tests-with-sd}
  \small

  \resizebox{0.95\linewidth}{!}{
  \begin{tabular}{lrrrrr}
    \toprule
    Dataset & PCS & PCO & HITON-PC & FSMB-PC & TM \\
    \midrule
    Insurance      
      & 6225.67 $\pm$ 71.00 
      & 4548.33 $\pm$ 131.23 
      & 3292.00 $\pm$ 100.58 
      & 4449.67 $\pm$ 75.80 
      & \textbf{219.33 $\pm$ 20.21} \\
    
    Water        
      & 1498.33 $\pm$ 50.82
      & 1256.33 $\pm$ 43.15
      & 1633.33 $\pm$ 6.66
      & 1766.67 $\pm$ 66.73
      & \textbf{200.33 $\pm$ 11.85} \\

    Alarm        
      & 4515.67 $\pm$ 199.83
      & 3150.33 $\pm$ 70.71
      & 3464.00 $\pm$ 28.58
      & 5548.67 $\pm$ 150.98
      & \textbf{271.33 $\pm$ 21.08} \\

    Barley      
      & 12638.00 $\pm$ 825.43
      & 9000.00 $\pm$ 737.76
      & 6874.67 $\pm$ 132.40
      & 9406.00 $\pm$ 39.60
      & \textbf{352.67 $\pm$ 15.82} \\

    Hailfinder   
      & 63461.00 $\pm$ 4753.65
      & 49519.00 $\pm$ 3256.31
      & 40346.67 $\pm$ 2029.87
      & 42877.67 $\pm$ 1819.43
      & \textbf{378.00 $\pm$ 13.00} \\

    Hepar2      
      & 11656.00 $\pm$ 1673.04
      & 9577.67 $\pm$ 925.24
      & 9852.33 $\pm$ 309.87
      & 11235.33 $\pm$ 699.22
      & \textbf{591.00 $\pm$ 13.89} \\

    Munin1     
      & 677727.5 $\pm$ 29757.17
      & 388738.00 $\pm$ 17648.02
      & 149114.33 $\pm$ 6281.22
      & 379516.33 $\pm$ 12217.55
      & \textbf{1427.67 $\pm$ 48.42} \\
    
    \bottomrule
  \end{tabular}
  }
\end{table}

\begin{table}
  \centering
  \caption{Precision of different algorithms across datasets (mean ± standard deviation)}
  \label{tab:precision}
  \footnotesize

  \resizebox{0.70\linewidth}{!}{
  \begin{tabular}{lccccc}
    \toprule
    Dataset & PCS & PCO & HITON-PC & FSMB-PC & TM \\
    \midrule
    Insurance       
      & \textbf{0.98 $\pm$ 0.03} 
      & 0.97 $\pm$ 0.02 
      & 0.84 $\pm$ 0.03 
      & 0.92 $\pm$ 0.05 
      & 0.94 $\pm$ 0.05 \\

    Water        
      & 0.98 $\pm$ 0.03
      & 0.97 $\pm$ 0.03
      & 0.83 $\pm$ 0.04
      & 0.88 $\pm$ 0.05
      & \textbf{1.0 $\pm$ 0.0} \\

    Alarm        
      & \textbf{0.99 $\pm$ 0.02}
      & \textbf{0.99 $\pm$ 0.02}
      & 0.82 $\pm$ 0.02
      & 0.91 $\pm$ 0.04
      & 0.94 $\pm$ 0.02 \\

    Barley      
      & \textbf{0.97 $\pm$ 0.03}
      & 0.95 $\pm$ 0.02
      & 0.74 $\pm$ 0.02
      & 0.76 $\pm$ 0.03
      & 0.84 $\pm$ 0.02 \\

    Hailfinder   
      & 0.48 $\pm$ 0.01
      & 0.48 $\pm$ 0.01
      & 0.45 $\pm$ 0.01
      & 0.47 $\pm$ 0.01
      & \textbf{0.68 $\pm$ 0.02} \\

    Hepar2      
      & \textbf{0.99 $\pm$ 0.02}
      & 0.98 $\pm$ 0.02
      & 0.9 $\pm$ 0.05
      & 0.95 $\pm$ 0.03
      & 0.98 $\pm$ 0.03 \\

    Munin1     
      & \textbf{0.80 $\pm$ 0.02}
      & 0.75 $\pm$ 0.02
      & 0.37 $\pm$ 0.02
      & 0.48 $\pm$ 0.02
      & 0.75 $\pm$ 0.02 \\
    
    \bottomrule
  \end{tabular}
  }
\end{table}

\begin{table}
  \centering
  \caption{Recall of different algorithms across datasets (mean ± standard deviation)}
  \label{tab:recall}
  \footnotesize

  \resizebox{0.70\linewidth}{!}{
  \begin{tabular}{lccccc}
    \toprule
    Dataset & PCS & PCO & HITON-PC & FSMB-PC & TM \\
    \midrule
    Insurance       
      & 0.56 $\pm$ 0.03 
      & 0.58 $\pm$ 0.03 
      & \textbf{0.81 $\pm$ 0.01} 
      & 0.77 $\pm$ 0.02 
      & 0.57 $\pm$ 0.02 \\

    Water        
      & 0.29 $\pm$ 0.02
      & 0.31 $\pm$ 0.03
      & \textbf{0.47 $\pm$ 0.03}
      & 0.41 $\pm$ 0.01
      & 0.31 $\pm$ 0.01 \\

    Alarm        
      & 0.8 $\pm$ 0.02
      & 0.75 $\pm$ 0.04
      & \textbf{0.94 $\pm$ 0.02}
      & \textbf{0.94 $\pm$ 0.01}
      & 0.81 $\pm$ 0.02 \\

    Barley      
      & 0.45 $\pm$ 0.06
      & 0.45 $\pm$ 0.04
      & \textbf{0.79 $\pm$ 0.02}
      & 0.74 $\pm$ 0.02
      & 0.4 $\pm$ 0.06 \\

    Hailfinder   
      & 0.66 $\pm$ 0.02
      & 0.66 $\pm$ 0.02
      & \textbf{0.92 $\pm$ 0.03}
      & 0.9 $\pm$ 0.02
      & 0.71 $\pm$ 0.01 \\

    Hepar2      
      & 0.33 $\pm$ 0.03
      & 0.34 $\pm$ 0.03
      & \textbf{0.68 $\pm$ 0.02}
      & 0.64 $\pm$ 0.01
      & 0.42 $\pm$ 0.02 \\

    Munin1     
      & 0.42 $\pm$ 0.01
      & 0.41 $\pm$ 0.02
      & \textbf{0.64 $\pm$ 0.01}
      & 0.59 $\pm$ 0.02
      & 0.42 $\pm$ 0.03 \\
    
    \bottomrule
  \end{tabular}
  }
\end{table}

\begin{table}
  \centering
  \caption{F1-score of different algorithms across datasets (mean ± standard deviation)}
  \label{tab:f1score}
  \footnotesize

  \resizebox{0.70\linewidth}{!}{
  \begin{tabular}{lccccc}
    \toprule
    Dataset & PCS & PCO & HITON-PC & FSMB-PC & TM \\
    \midrule
    Insurance       
      & 0.71 $\pm$ 0.02 
      & 0.73 $\pm$ 0.03 
      & 0.82 $\pm$ 0.02 
      & \textbf{0.84 $\pm$ 0.03} 
      & 0.71 $\pm$ 0.02 \\

    Water        
      & 0.45 $\pm$ 0.02
      & 0.47 $\pm$ 0.03
      & \textbf{0.60 $\pm$ 0.03}
      & 0.57 $\pm$ 0.02
      & 0.47 $\pm$ 0.01 \\

    Alarm        
      & 0.90 $\pm$ 0.04
      & 0.86 $\pm$ 0.02
      & 0.88 $\pm$ 0.02
      & \textbf{0.93 $\pm$ 0.03}
      & 0.87 $\pm$ 0.02 \\

    Barley      
      & 0.61 $\pm$ 0.06
      & 0.61 $\pm$ 0.05
      & 0.75 $\pm$ 0.01
      & \textbf{0.76 $\pm$ 0.02}
      & 0.55 $\pm$ 0.06 \\

    Hailfinder   
      & 0.56 $\pm$ 0.01
      & 0.55 $\pm$ 0.01
      & 0.60 $\pm$ 0.02
      & 0.62 $\pm$ 0.01
      & \textbf{0.70 $\pm$ 0.01} \\

    Hepar2      
      & 0.49 $\pm$ 0.03
      & 0.50 $\pm$ 0.03
      & \textbf{0.78 $\pm$ 0.03}
      & 0.76 $\pm$ 0.02
      & 0.59 $\pm$ 0.02 \\

    Munin1     
      & \textbf{0.55 $\pm$ 0.01}
      & 0.53 $\pm$ 0.02
      & 0.47 $\pm$ 0.02
      & 0.53 $\pm$ 0.01
      & 0.54 $\pm$ 0.03 \\
    
    \bottomrule
  \end{tabular}
  }
\end{table}

As shown in Table~\ref{tab:ci-tests-with-sd}, the TM-based method consistently required far fewer CI tests compared to traditional algorithms such as PCS, PCO, HITON-PC, and FSMB-PC. For instance, in the \textit{Munin1} network, TM performed only 1,428 CI tests on average, whereas PCS and PCO needed 6,77,727 and 3,88,738 tests, respectively. This demonstrates a substantial reduction in computational effort, highlighting the efficiency and scalability of TM, especially for large and complex networks.

Regarding predictive performance (Tables~\ref{tab:precision}--\ref{tab:recall}), the TM-based approach showed strong precision and acceptable recall across networks of varying sizes. For smaller networks such as \textit{Insurance} and \textit{Water}, TM achieved high precision (0.94 and 1.00, respectively), comparable to or even exceeding PCS and PCO, although recall remained moderate (0.57 and 0.31). This reflects the usual trade-off between accurately identified edges and completeness.

On medium-sized networks such as \textit{Alarm} and \textit{Barley}, TM maintained a generally balanced performance, with precision between 0.94--0.84 and recall between 0.81--0.40. This indicates reliable edge identification while keeping false positives under control.

For larger networks such as \textit{Hailfinder}, \textit{Hepar2}, and \textit{Munin1}, TM preserved competitive precision (0.68--0.98) while maintaining reasonable recall. Notably, TM achieved a recall of 0.71 on \textit{Hailfinder}---higher than the initially stated range---along with 0.42 on \textit{Hepar2} and 0.42 on \textit{Munin1}. These results demonstrate the algorithm's ability to recover a substantial portion of true dependencies even in highly complex topologies.

In terms of F1-score (Tables~\ref{tab:f1score}), which balances precision and recall, TM remained competitive with the benchmark methods. While FSMB and HITON-PC occasionally outperformed TM on smaller networks such as \textit{Alarm} (F1 = 0.93) and \textit{Barley} (F1 = 0.76), TM showed comparable or stronger performance on larger networks. Notably, TM achieved the highest F1-score on \textit{Hailfinder} (0.70) and performed competitively on \textit{Munin1} (0.54), indicating robustness in reconstructing the underlying network structure.

Overall, the results show that the precision of the TM-based method is competitive with PCS and PCO, and is generally higher than that of FSMB-PC and HITON-PC across most datasets. In contrast, the recall of FSMB-PC and HITON-PC is consistently higher than that of TM, as well as higher than the recall achieved by PCS and PCO. This highlights a clear trade-off: TM offers strong precision with very low computational cost, while FSMB-PC and HITON-PC prioritise higher recall at the expense of performing a much larger number of CI tests.

\section{Conclusion} \label{Conclusion}

In this paper, we proposed a TM-based Parent-Children algorithm for efficient Bayesian network learning, which identifies the most influential variables and performs a small, fixed number of conditional independence tests, resulting in linear complexity with respect to the number of features. Experimental results show that the method dramatically reduces computational cost while maintaining competitive precision, recall, and F1-scores across networks of varying sizes, demonstrating its scalability and effectiveness for categorical datasets. A key limitation is that by considering only the top-2 variables per target, the approach may miss dependencies in networks with larger Markov blankets. Future work should focus on extending the method to adaptively consider more variables per target, aiming for infinite-sample assumption and broader applicability without sacrificing efficiency.

\bibliographystyle{plainnat} 
\bibliography{reference} 

\appendix
\section{Hyperparameter tuning}

To optimise the performance of the proposed TM-based technique, we carried out hyperparameter tuning using the \textit{Insurance} and \textit{Alarm} Bayesian Networks as representative examples. The three main hyperparameters investigated were the number of clauses ($B$) and one tuning parameter called Specificity ($S$). We tested three values of $B$ (10, 50, and 100) to examine the effect of increasing model complexity, and three values of $S$ (1, 5, and 10) to study the impact of specificity on performance.

\begin{figure}[htbp]
  \centering
  \includegraphics[width=0.90\linewidth]{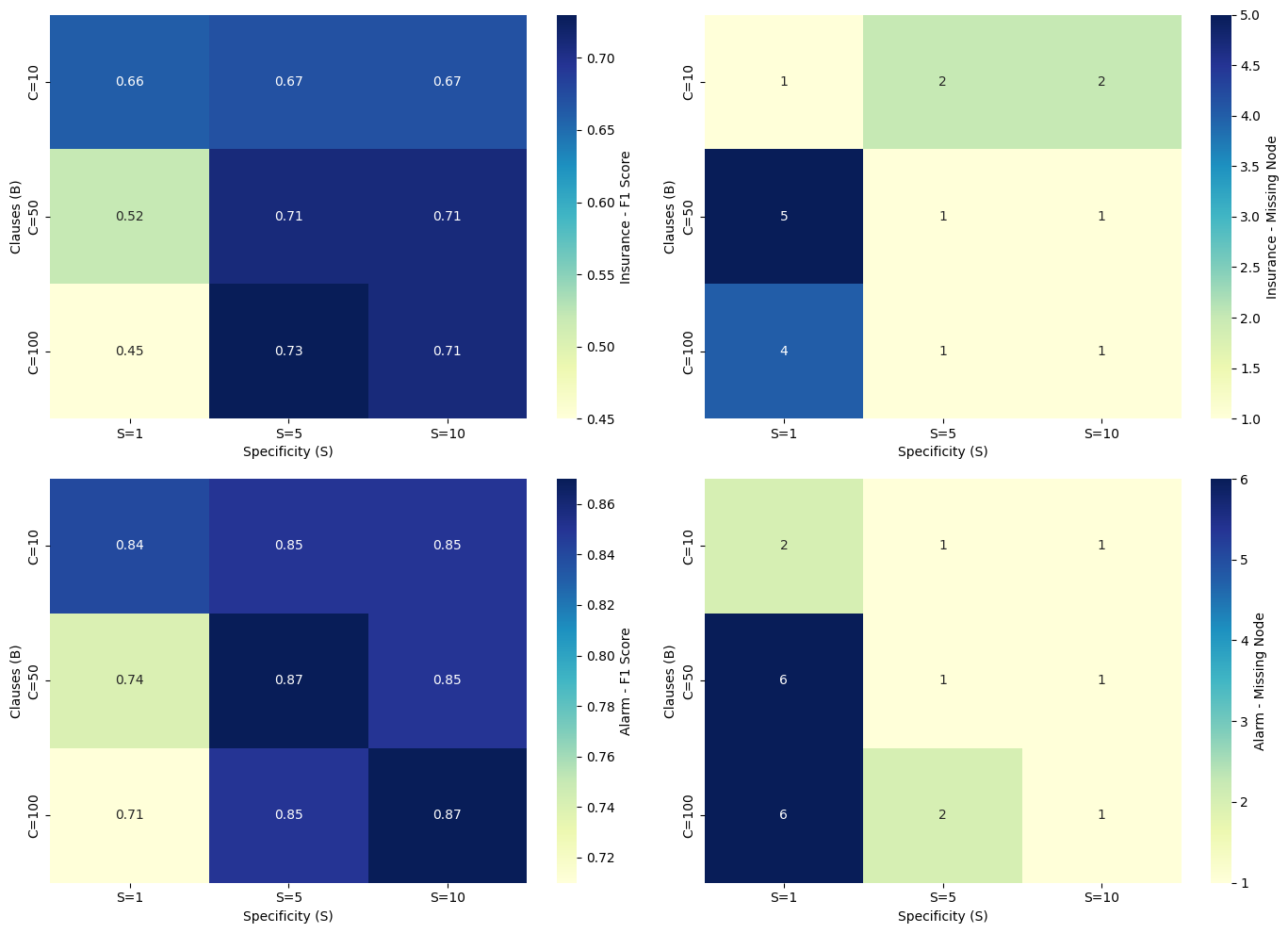} 
  \caption{Hyperparameter tuning.}
  \label{fig:heatmap}
\end{figure}

The results are summarised in Figure~\ref{fig:heatmap} using heatmaps for both the F1 score and the number of missing nodes. Each heatmap illustrates how the combination of $B$ (rows) and $S$ (columns) affects the respective metric. For the \textit{Insurance} network, higher F1 scores were generally observed for larger clause counts ($B = 50$ or $100$) and higher specificity ($S = 5$ or $10$), indicating that increasing model capacity improves the recovery of network structure. Simultaneously, missing nodes decreased for larger $B$ and moderate-to-high $S$ values, suggesting better coverage of the true network variables.

For the \textit{Alarm} network, F1 scores were consistently high across most hyperparameter combinations, though slightly lower for the smallest $B$ and $S$ values. Missing nodes reduced as $B$ and $S$ increased, confirming that both parameters contribute to more accurate network reconstruction.

\section{More results}

To supplement the results reported in the main text, we present additional structural accuracy measures in the  (Tables~\ref{tab:missing_nodes}--\ref{tab:shd}), focusing on Structural Hamming Distance (SHD) as well as its components: missing nodes, missing edges, and wrong edges. The SHD provides a comprehensive measure of network accuracy, taking into account absent edges, extra edges. Lower values indicate structures closer to the true network. TM consistently achieved competitive SHD scores across datasets.

For instance, in the \textit{Insurance} network, TM recorded an SHD of $24 \pm 2$, comparable to PCS ($23 \pm 2$) and PCO ($22 \pm 2$), and slightly higher than FSMB-PC ($15 \pm 3$). In terms of missing nodes, TM missed only $1 \pm 1$, comparable to FSMB-PC ($0 \pm 0$) and better than PCS and PCO ($2 \pm 1$ and $2 \pm 0$, respectively). Regarding missing edges (Table~\ref{tab:missing_edges}), TM had $22 \pm 1$, comparable to PCS ($23 \pm 2$) and PCO ($22 \pm 2$), and for wrong edges (Table~\ref{tab:wrong_edges}), TM recorded $2 \pm 2$, which was among the lowest across methods.

In the \textit{Alarm} network, TM achieved an SHD of $11 \pm 2$, slightly higher than FSMB ($6 \pm 3$) and similar to PCO ($11 \pm 1$) and PCS ($9 \pm 1$). Missing nodes for TM were minimal ($0 \pm 0$), while missing edges ($9 \pm 1$) and wrong edges ($2 \pm 1$) remained low, indicating accurate edge recovery.

For larger networks such as \textit{Hailfinder}, TM demonstrated strong performance with an SHD of $40 \pm 2$, substantially outperforming PCS ($68 \pm 2$), PCO ($69 \pm 1$), HITON-PC ($79 \pm 3$) and FSMB-PC($72 \pm 2$). Missing nodes ($0 \pm 0$), missing edges ($16 \pm 4$), and wrong edges ($22 \pm 2$) further confirm TM’s robustness in recovering large network structures. Even for very large networks like \textit{Munin1}, TM maintained an SHD of $196 \pm 8$, with manageable missing nodes ($44 \pm 4$), missing edges ($158 \pm 8$), and wrong edges ($38 \pm 1$), compared to other methods such as HITON-PC ($389 \pm 20$ SHD, $291 \pm 20$ wrong edges) and FSMB ($278 \pm 10$ SHD, $168 \pm 13$ wrong edges).

Overall, across datasets, the TM-based method showed missing edges that were similar to or lower than PCS and PCO, but slightly higher than HITON-PC and FSMB. Wrong edges were consistently much lower than FSMB and HITON-PC, and comparable to PCS and PCO. These results reinforce that TM not only reduces computational effort substantially but also maintains strong structural accuracy, effectively recovering edges while minimizing both missing and incorrect elements.

\begin{table}[htbp]
  \centering
  \caption{Missing nodes across datasets (mean ± standard deviation)}
  \label{tab:missing_nodes}
  \footnotesize
  \resizebox{0.70\linewidth}{!}{
  \begin{tabular}{lccccc}
    \toprule
    Dataset & PCS & PCO & HITON-PC & FSMB-PC & TM \\
    \midrule
    Insurance      
      & 2 $\pm$ 1
      & 2 $\pm$ 0
      & \textbf{0 $\pm$ 0}
      & \textbf{0 $\pm$ 0}
      & 1 $\pm$ 1 \\

    Water        
      & \textbf{6 $\pm$ 0}
      & \textbf{6 $\pm$ 0}
      & \textbf{6 $\pm$ 0}
      & \textbf{6 $\pm$ 0}
      & \textbf{6 $\pm$ 0} \\

    Alarm        
      & 1 $\pm$ 0
      & 2 $\pm$ 1
      & 0 $\pm$ 1
      & 1 $\pm$ 1
      & \textbf{0 $\pm$ 0} \\

    Barley      
      & 5 $\pm$ 2
      & 5 $\pm$ 2
      & 1 $\pm$ 1
      & \textbf{0 $\pm$ 1} 
      & \textbf{0 $\pm$ 1} \\

    Hailfinder   
      & 0 $\pm$ 1
      & 0 $\pm$ 1
      & \textbf{0 $\pm$ 0}
      & \textbf{0 $\pm$ 0}
      & \textbf{0 $\pm$ 0} \\

    Hepar2      
      & 19 $\pm$ 3
      & 18 $\pm$ 3
      & \textbf{2 $\pm$ 2}
      & \textbf{2 $\pm$ 2}
      & 14 $\pm$ 1 \\

    Munin1     
      & 48 $\pm$ 5
      & 52 $\pm$ 2
      & \textbf{34 $\pm$ 0}
      & 34 $\pm$ 1
      & 44 $\pm$ 4 \\
    
    \bottomrule
  \end{tabular}
  }
\end{table}

\begin{table}[htbp]
  \centering
  \caption{Missing edges across datasets (mean ± standard deviation)}
  \label{tab:missing_edges}
  \footnotesize
  \resizebox{0.75\linewidth}{!}{
  \begin{tabular}{lccccc}
    \toprule
    Dataset & PCS & PCO & HITON-PC & FSMB-PC & TM \\
    \midrule
    Insurance      
      & 23 $\pm$ 2
      & 22 $\pm$ 2
      & \textbf{10 $\pm$ 1}
      & 12 $\pm$ 1
      & 22 $\pm$ 1 \\

    Water        
      & 46 $\pm$ 1
      & 45 $\pm$ 2
      & \textbf{35 $\pm$ 2}
      & 38 $\pm$ 1
      & 45 $\pm$ 1 \\

    Alarm        
      & 9 $\pm$ 1
      & 11 $\pm$ 2
      & \textbf{2 $\pm$ 1}
      & \textbf{2 $\pm$ 1}
      & 9 $\pm$ 1 \\

    Barley      
      & 46 $\pm$ 5
      & 46 $\pm$ 4
      & \textbf{17 $\pm$ 2}
      & 21 $\pm$ 1
      & 50 $\pm$ 5 \\

    Hailfinder   
      & 22 $\pm$ 1
      & 22 $\pm$ 1
      & \textbf{5 $\pm$ 2}
      & 6 $\pm$ 1
      & 16 $\pm$ 4 \\

    Hepar2      
      & 82 $\pm$ 3
      & 81 $\pm$ 3
      & \textbf{38 $\pm$ 3}
      & 43 $\pm$ 1
      & 71 $\pm$ 2 \\

    Munin1     
      & 156 $\pm$ 4
      & 160 $\pm$ 4
      & \textbf{98 $\pm$ 3}
      & 111 $\pm$ 4
      & 158 $\pm$ 8 \\
    
    \bottomrule
  \end{tabular}
  }
\end{table}

\begin{table}[htbp]
  \centering
  \caption{Wrong edges across datasets (mean ± standard deviation)}
  \label{tab:wrong_edges}
  \footnotesize
  \resizebox{0.75\linewidth}{!}{
  \begin{tabular}{lccccc}
    \toprule
    Dataset & PCS & PCO & HITON-PC & FSMB-PC & TM \\
    \midrule
    Insurance      
      & \textbf{0 $\pm$ 1}
      & 1 $\pm$ 1
      & 8 $\pm$ 2
      & 3 $\pm$ 2
      & 2 $\pm$ 2 \\

    Water        
      & 0 $\pm$ 1
      & 1 $\pm$ 1
      & 6 $\pm$ 2
      & 4 $\pm$ 2
      & \textbf{0 $\pm$ 0} \\

    Alarm        
      & \textbf{0 $\pm$ 1}
      & \textbf{0 $\pm$ 1}
      & 9 $\pm$ 1
      & 4 $\pm$ 2
      & 2 $\pm$ 1 \\

    Barley      
      & \textbf{1 $\pm$ 1}
      & 2 $\pm$ 1
      & 23 $\pm$ 2
      & 20 $\pm$ 2
      & 6 $\pm$ 1 \\

    Hailfinder   
      & 46 $\pm$ 2
      & 47 $\pm$ 2
      & 74 $\pm$ 2
      & 66 $\pm$ 2
      & \textbf{22 $\pm$ 2} \\

    Hepar2      
      & \textbf{0 $\pm$ 1}
      & \textbf{0 $\pm$ 1}
      & 9 $\pm$ 5
      & 4 $\pm$ 2
      & 1 $\pm$ 1 \\

    Munin1     
      & \textbf{28 $\pm$ 3}
      & 37 $\pm$ 3
      & 291 $\pm$ 20
      & 168 $\pm$ 13
      & 38 $\pm$ 1 \\
    
    \bottomrule
  \end{tabular}
  }
\end{table}

\begin{table}[htbp]
  \centering
  \caption{Structural hamming distance across datasets (mean ± standard deviation)}
  \label{tab:shd}
  \footnotesize

  \resizebox{0.75\linewidth}{!}{
  \begin{tabular}{lccccc}
    \toprule
    Dataset & PCS & PCO & HITON-PC & FSMB-PC & TM \\
    \midrule
    Insurance      
      & 23 $\pm$ 2
      & 22 $\pm$ 2
      & 18 $\pm$ 2
      & \textbf{15 $\pm$ 3}
      & 24 $\pm$ 2 \\

    Water        
      & 47 $\pm$ 1
      & 46 $\pm$ 2
      & \textbf{41 $\pm$ 3}
      & 42 $\pm$ 2
      & 45 $\pm$ 1 \\

    Alarm        
      & 9 $\pm$ 1
      & 11 $\pm$ 1
      & 11 $\pm$ 2
      & \textbf{6 $\pm$ 3}
      & 11 $\pm$ 2 \\

    Barley      
      & 47 $\pm$ 5
      & 48 $\pm$ 4
      & \textbf{40 $\pm$ 2}
      & 40 $\pm$ 4
      & 56 $\pm$ 4 \\

    Hailfinder   
      & 68 $\pm$ 2
      & 69 $\pm$ 1
      & 79 $\pm$ 3
      & 72 $\pm$ 2
      & \textbf{40 $\pm$ 2} \\

    Hepar2      
      & 82 $\pm$ 3
      & 81 $\pm$ 3
      & 47 $\pm$ 7
      & \textbf{47 $\pm$ 3}
      & 71 $\pm$ 2 \\

    Munin1     
      & \textbf{184 $\pm$ 6}
      & 197 $\pm$ 6
      & 389 $\pm$ 20
      & 278 $\pm$ 10
      & 196 $\pm$ 8 \\
    
    \bottomrule
  \end{tabular}
  }
\end{table}

\end{document}